\documentclass{article}
\pdfoutput=1
\usepackage{spconf,amsmath,graphicx}
\usepackage{graphicx}
\usepackage{booktabs}
\usepackage{comment}
\usepackage{caption}
\usepackage[pagebackref=true,breaklinks=true,colorlinks,bookmarks=false]{hyperref}

\newcommand{\fig}[1]{Fig.~\ref{fig:#1}}
\newcommand{\tab}[1]{Table~\ref{tab:#1}}

\usepackage{graphicx}
\usepackage{amsmath}
\usepackage{amssymb}
\usepackage{booktabs}

\usepackage{xspace}

\title{Evaluating Context for Deep Object Detectors}

\name{Osman Semih Kayhan
\qquad Jan C. van Gemert}
\address{Computer Vision Lab, Delft University of Technology}

\begin{document}
%
\makeatletter
\let\@oldmaketitle\@maketitle
\renewcommand{\@maketitle}{\@oldmaketitle
  \includegraphics[width=\linewidth,]
    {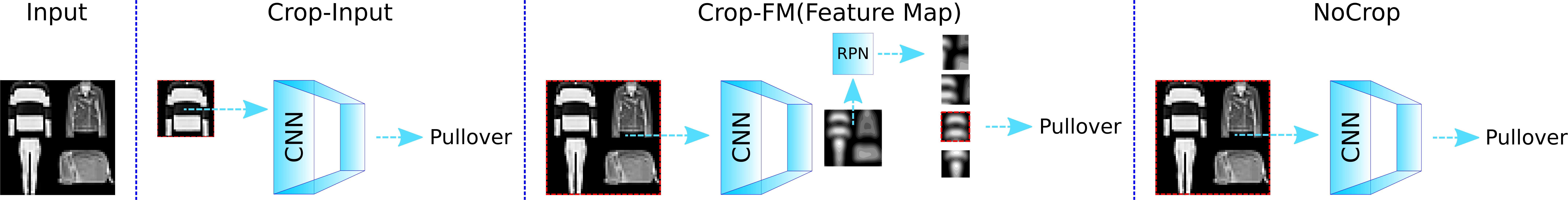}
    \captionof{figure}{ROI handling in deep object detectors. Crop-Input (no context) crops the ROI from the input image before the CNN. Crop-FM (partial context) inputs the full image and crops the ROIs from the featuremap. NoCrop (full context) has a single-stage and uses the full image as input to the CNN.}\label{fig:toy_archs}
    \bigskip}
\makeatother

\maketitle

\begin{abstract}

Which object detector is suitable for your context sensitive task? 
Deep object detectors exploit scene context for recognition differently. In this paper, we group object detectors into 3 categories in terms of context use: \textit{no context} by cropping the input (RCNN), \textit{partial context} by cropping the featuremap (two-stage methods) and \textit{full context} without any cropping (single-stage methods). We systematically evaluate the effect of context for each deep detector category. We create a fully controlled dataset for varying context and investigate the context for deep detectors. We also evaluate gradually removing the background context and the foreground object on MS COCO. 
We demonstrate that single-stage and two-stage object detectors can and will use the context by virtue of their large receptive field. Thus, choosing the best object detector may depend on the application context.
\footnote{\url{https://github.com/oskyhn/Detectors-Context}}

\end{abstract}

\section{Introduction}

Objects are rarely photographed alone. An image may contain several other objects or varying scene background. This background context may correlate with the object and thus possibly exploited by a learned object detector such as detecting a chair next to a table and missing the same chair in a football field. 
For some applications, the context should not matter like an object placed in a box of a retail system. For other applications, the context is an important cue, such as images from an MRI scanner. In this paper we evaluate the effect of context on popular deep object detectors.

For deep object detectors, there are works on the
context around the object~\cite{ gidaris2015object}, as scene context~\cite{gupta2015exploring, Sun_2017_CVPR, Liu_2018}, as combinations of local parts and the global image structure~\cite{zhu2017couplenet, fan2018object}, by using multi-scale fusion and attention \cite{lim2019small}, and by using recurrent neural networks~\cite{bell2016inside}.
Besides, to have a robust decision, \cite{zhang2019bag} uses mixup augmentation and \cite{singh2020don} investigates how to disentangle object from its co-occurring context.
\cite{zhang2019putting} investigates the contextual effect on visual recognition with various ways and compare with human performance.
\cite{Barnea_2019_CVPR} explores the bounds of object classes by using contextual information and show the cases when is beneficial to use or discard. 
\cite{kayhan2021hallucination} indicates that contextual information can make the detectors hallucinate some objects and their locations whilst the objects are not even in the images. 
In addition, \cite{kayhan2020translation, manfredi2020shift} show absolute spatial position of object can affect the performance.
Differently,
we classify object detectors in terms of context use and analyze the effect of context for these detector types.

Popular deep object detectors follow either a single-stage or a two-stage approach. Two-stage detectors consist of class-agnostic region proposals and detection parts. RCNN~\cite{girshick2015region} is the earliest deep two-stage detector that crops the ROIs from the input image before feeding the CNN backbone, without accessing context.  
Faster RCNN \cite{ren2015faster} introduces the trainable Region Proposal Network (RPN) and each candidate region is cropped from deep featuremap. 
Faster RCNN is the most common two-stage detector and a great inspiration to other detectors~\cite{zhu2017couplenet, fan2018object, lin2017feature, dai2017deformable},
hence we evaluate Faster-RCNN as the prototypical two-stage detector.
Single-stage detectors do not use any proposal method and obtain the detection in a single run. 
YOLO \cite{redmon2016you} treats detection as a regression problem and detects the objects
from a full image. 
In YOLOv2 \cite{Redmon_2017_CVPR} and YOLOv3 \cite{redmon2018yolov3}, the method is improved by using a deeper backbone model, multi-scale training, high resolution input and anchor boxes. 
SSD \cite{liu2016ssd} predicts category scores and box offsets for a fixed set of anchors from different scales.
RetinaNet \cite{lin2017focal} proposes focal loss which focuses on the hard training samples and converges faster.
EfficientDet \cite{tan2019efficientdet} scales the model and proposes fast multi-scale feature fusion. 
For our experiments, we choose YOLOv3 since it uses anchor boxes and multi-scale training, thus comparable with Faster RCNN.

In this paper, we assume that context is formed as everything around the object including other objects.
We classify object detectors on how they use context (Fig.~\ref{fig:toy_archs}): (i) no context (RCNN), (ii) partial context (Faster RCNN) and (iii) full context (YOLO)  on a fully-controlled dataset, see Fig.~\ref{fig:toy_dataset}. 
Using context is beneficial if the object correlates with its environment. However, the performance is reduced when the context is incoherent. 
We demonstrate the effect of context on detector performance by increasing the context around to object in each testing case. Also, we indicate how much contribution the detector can obtain by only using the context information  without a visible main object.

We have the following contributions: 
First, we show that modern deep object detectors can access context by virtue of their receptive field size even if the object regions are cropped from the featuremap.  Second, the effect of context is evaluated quantitatively on most common object detection networks. To conclude, we indicate that single and two stages networks employ contextual information except methods crop the ROIs from the input such as RCNN.

\section{Experimental evaluation of context}
We analyze the effects of various contextual correlations for common detector types on a fully-controlled context dataset and evaluate context with natural images.

We categorize deep object detectors (see~\fig{toy_archs}) as:

\textbf{Crop-Input}. We base this class of detectors on the seminal RCNN~\cite{girshick2015region} approach, which originally
uses class-agnostic object proposal bounding boxes~\cite{uijlings2013selective} 
which are cropped from the input and fed to a CNN backbone for feature extraction, then the extracted features are used for detection.
Since the method crops the proposals from the input image before feature extraction, it does not access any context beyond the bounding box, thus we call it 'no context' method. In reality, a network may retrieve minimal context between an object and the area inside the bounding box. 

\textbf{Crop-FM}. This class of detectors crops bounding boxes from CNN featuremaps. The seminal example is based on Faster RCNN~\cite{ren2015faster} which has two stages: a detection head for object classification and an 
RPN which outputs candidate object boxes. These boxes are cropped by ROI pooling from featuremaps. These featuremaps are deep in the network and thus are the result of convolutions with a large receptive field.
Due to such large receptive fields, the featuremap crops include context information beyond the cropped regions which can be exploited for recognition.

\textbf{NoCrop}. This class of object detectors does not crop at all and includes most of the single-stage object detectors such as YOLO detectors~\cite{redmon2016you,Redmon_2017_CVPR,redmon2018yolov3}.
Predictions are made by using the full featuremap and thus can exploit all context.

\subsection{Evaluating object-context correlations}
\begin{figure}[t]
    \centering
        \includegraphics[width=0.99\linewidth]{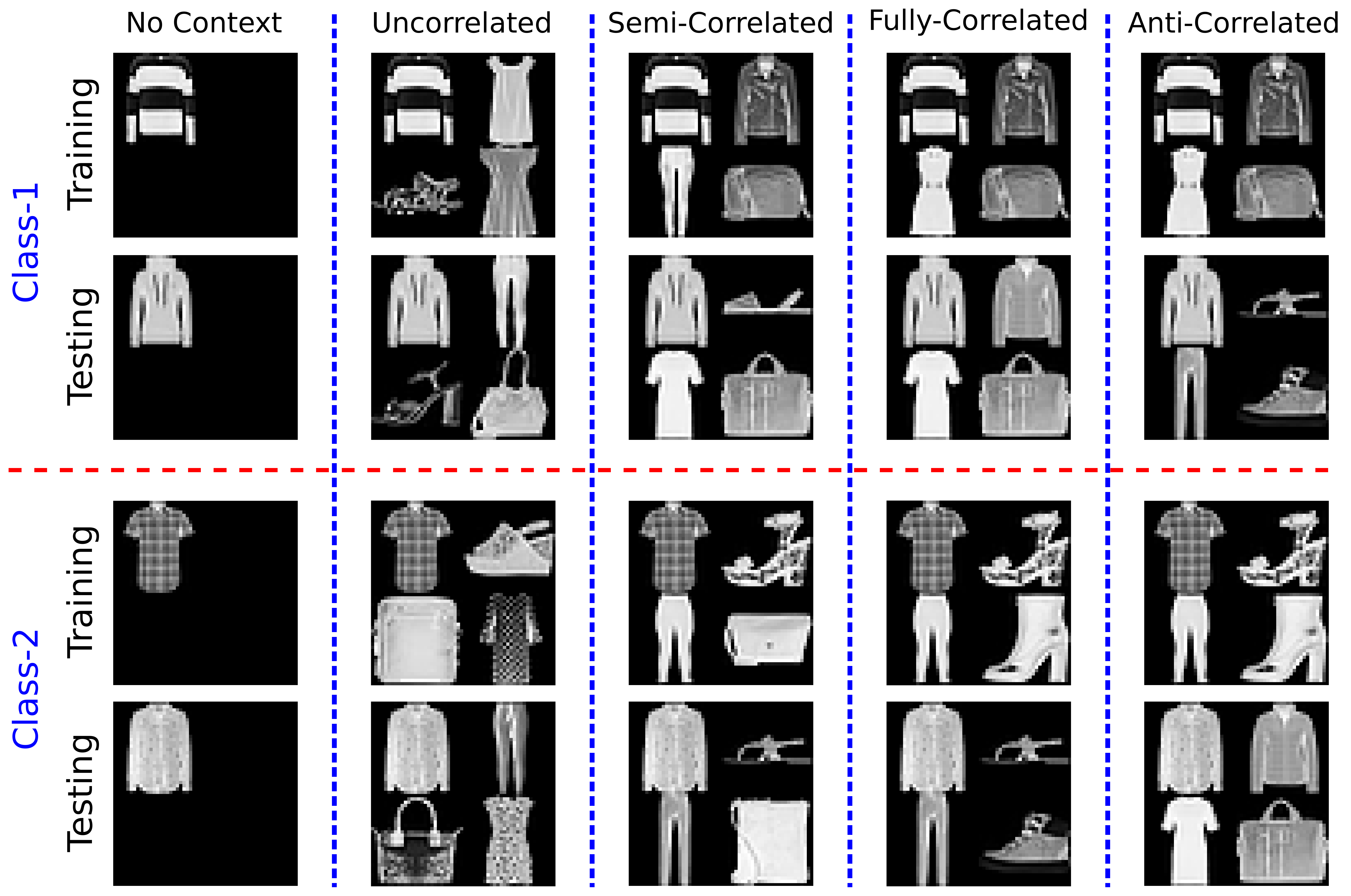}
    \caption{Q-FMNIST training and testing samples from each class. Controlled setting where the main object (top left) is surrounded by a varying degree of object-context correlations.}
    \label{fig:toy_dataset}
\end{figure}

It is difficult to vary the correlation between an object and its context in real images. Thus, we create a fully controlled context-sensitive dataset from the 10-class Fashion MNIST~\cite{xiao2017fashion}. To vary object-context correlation we create 6,000 images (2000 per training, validating and test set) to form the Quadrant-FMNIST (Q-FMNIST) dataset by placing images in quadrants. We create a 2-class object detection problem where the top-left quadrant has the object of interest (class-1: 'Pullover' and class-2: 'Shirt') and the other 3 quadrants are filled with images of other 8 classes which is how we vary object-context correlation.
Namely, these 8 classes become background for each image. We have 5 degrees of object-context correlations, shown in~\fig{toy_dataset}.

\textbf{No context:} The background is full black.

\textbf{Uncorrelated context:} The 3 locations of the template are filled randomly from the 8 context classes.

\textbf{Semi-correlated context:} Up-right corner, bottom-left corner and bottom-right corner locations of class-1 are filled with respectively 'Dress', 'Coat' and 'Bag' classes (common context for class-1) by the probability of $75\%$ and with 'Trouser', 'Sandal' and 'Ankle bot' classes by $25\%$. Similarly, class-2 images are filled with 'Trouser', 'Sandal' and 'Ankle bot' classes (common context for class-2) by $75\%$ and 'Dress', 'Coat' and 'Bag' classes by $25\%$.

\textbf{Fully-correlated context:} This mode represents the cases when the objects are observed always in similar context, such as cows and horses are together on a grass field. The context is placed in a structured way in terms of location, the same context occurs in the same place in the train and test set. For class-1 images, coat, dress and bag objects are placed respectively up-right corner, bottom-left corner and bottom-right corner. Likewise, For class-2, sandal, trouser and ankle bot objects are placed with the same order as the coat, dress and bag objects. The context objects are not alternated.

\textbf{Anti-correlated context:} Training set is built by using fully-structured context train set, however, testing is done by filling with incoherent context by switching the class-specific context. This mode illustrates the cases when the object is seen in an unusual context, such as a shoe in a plate. Class-1 images are filled with 'Trouser', 'Sandal' and 'Ankle bot' classes and likewise class-2 images are filled with  'Dress', 'Coat' and 'Bag' classes respectively up-right corner, bottom-left corner and bottom-right corner. Namely, the context of class-1 and class-2 are swapped.

\begin{table}
    \centering
	\renewcommand{\arraystretch}{0.7}
	\label{tab:toy_results}
	\begin{tabular}{@{}llll@{}}
		\toprule
		\toprule
		Context & Crop-Input     & Crop-FM    & NoCrop     \\ \midrule
		No context             & $87.9 \pm 1.0$ & $\bf{88.1 \pm 0.4}$ & $86.9 \pm 0.7$ \\
		Uncorrelated            & $\bf{87.9 \pm 1.0}$ & $86.7 \pm 0.7$ & $83.7 \pm 1.1$ \\
		Semi-correlated  & $87.9 \pm 1.0$ & $89.3 \pm 0.8$ & $\bf{90.4 \pm 0.5}$ \\
		Fully-correlated & $87.9 \pm 1.0$ & $99.7 \pm 0.1$ & $\bf{100 \pm 0.0}$  \\
		Anti-correlated       & $\bf{87.9 \pm 1.0}$ & $1.8 \pm 0.2$ & $0.0  \pm  0.0 $ \\ \bottomrule
	\end{tabular}
	\caption{Accuracy on Q-FMNIST. Context affects Crop-FM and NoCrop detectors. For correlated context results improve. Uncorrelated context is worse than no context. Anti-correlated context is detrimental.  }%
  \label{tab:toy_results}%
\end{table}

We instantiate each of the three  deep object detector classes in~\fig{toy_archs} with two convolution layers with 6 and 16 3x3 filters, two max pooling layers, one fully connected layer with 128 neurons and softmax classifier. We use the ground truth location to crop bounding boxes for the Crop-Input model. For the Crop-FM model, RoiAlign~\cite{he2017maskRCNN} 
is used for cropping ROIs from the featuremap. Each method is trained 5 times for 15 epochs with the AdaDelta optimizer.

\textbf{Results.} In~\tab{toy_results}, the Crop-Input detector disregards all context. For the Crop-FM and NoCrop detectors, the 'no context' setting gives an object-only baseline. Adding 'semi-structured context' improves, and adding 'fully correlated context' even more. Interestingly, adding 'uncorrelated context' decreases results, whereas 'anti-correlated context' completely misclassifies the objects. As the NoCrop detector uses the full image, it is more sensitive to context changes than the Crop-FM. Being more sensitive can be an advantage for correlated object-contexts, yet can be detrimental for random context or when an object is placed outside the usual context where the context may outweigh the object itself.

\subsection{Evaluating context on natural images}

\begin{figure}
	\includegraphics[width=0.99\linewidth]{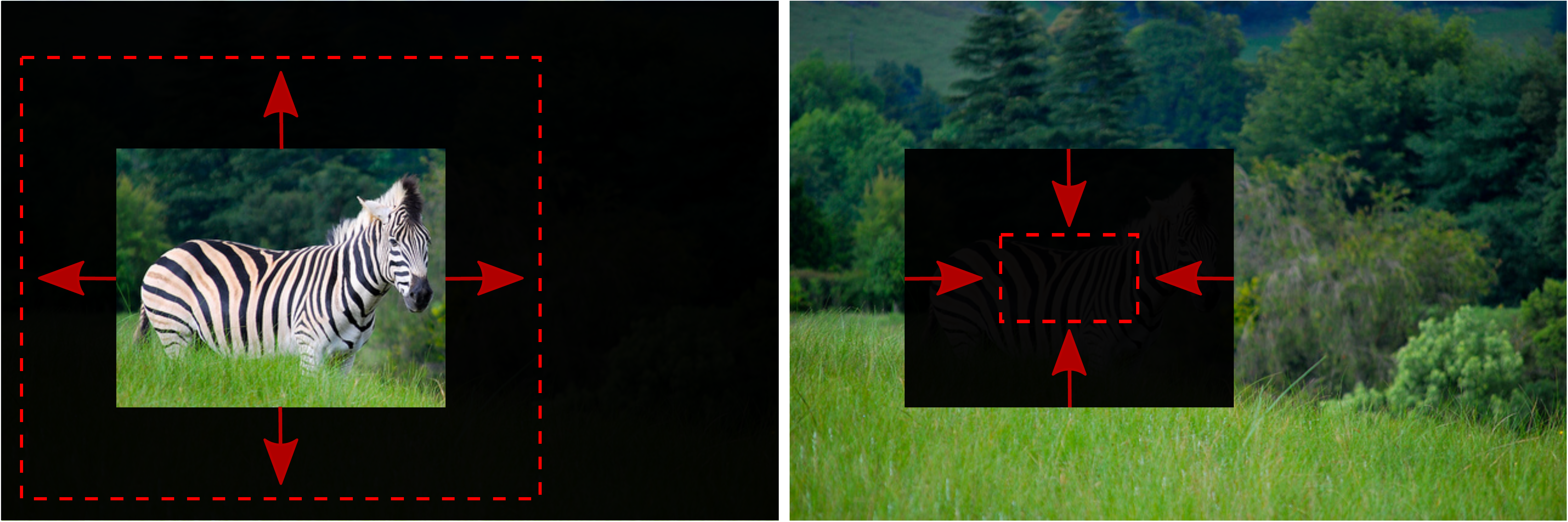}
	\caption{Example of hiding background (left) and hiding foreground (right). Hiding BG incrementally adds  background pixels. Hiding FG incrementally adds  object pixels.}
	\label{fig:bg_fg}
\end{figure}

We investigate the context for natural images on the COCO minival 2014 split~\cite{lin2014microsoft}. We evaluate two settings: \textit{Hiding background} and \textit{Hiding foreground}, see~\fig{bg_fg}. We begin with the ground truth object bounding box. For \textit{hiding BG}, we start without any context using a black background and incrementally add more background pixels on each side of the object. For \textit{hiding FG}, we start without an object and make the bounding box black and incrementally add object pixels towards the center on each side. We  increase pixels in the range  $\in \{0,5,10,25,50,100,150,200,250,..\}$  until reaching the full image for both settings. If the context of an image reached an image border on one side, it stops there, yet, the change continues on the other object sides. 

For Crop-Input we use a variation of R-CNN~\cite{girshick2015region} where we use a softmax classifier based on an ImageNet pretrained Alexnet. The object crops are resized as 227x227 and trained from scratch 35 epochs with SGD  for an initial learning rate of 1e-3. 
For Crop-FM, we use Faster RCNN with ROI Align to crop ROIs from the featuremap. An existing COCO pretrained network is used with a Resnet-50 backbone with an FPN~\cite{lin2017feature}. For NoCrop, we use YOLO version 3 \cite{redmon2018yolov3} which is fully-convolutional and has 75 layers with skip connections. 
\begin{figure*}[t]
	\centering
	\includegraphics[width=0.78\linewidth]{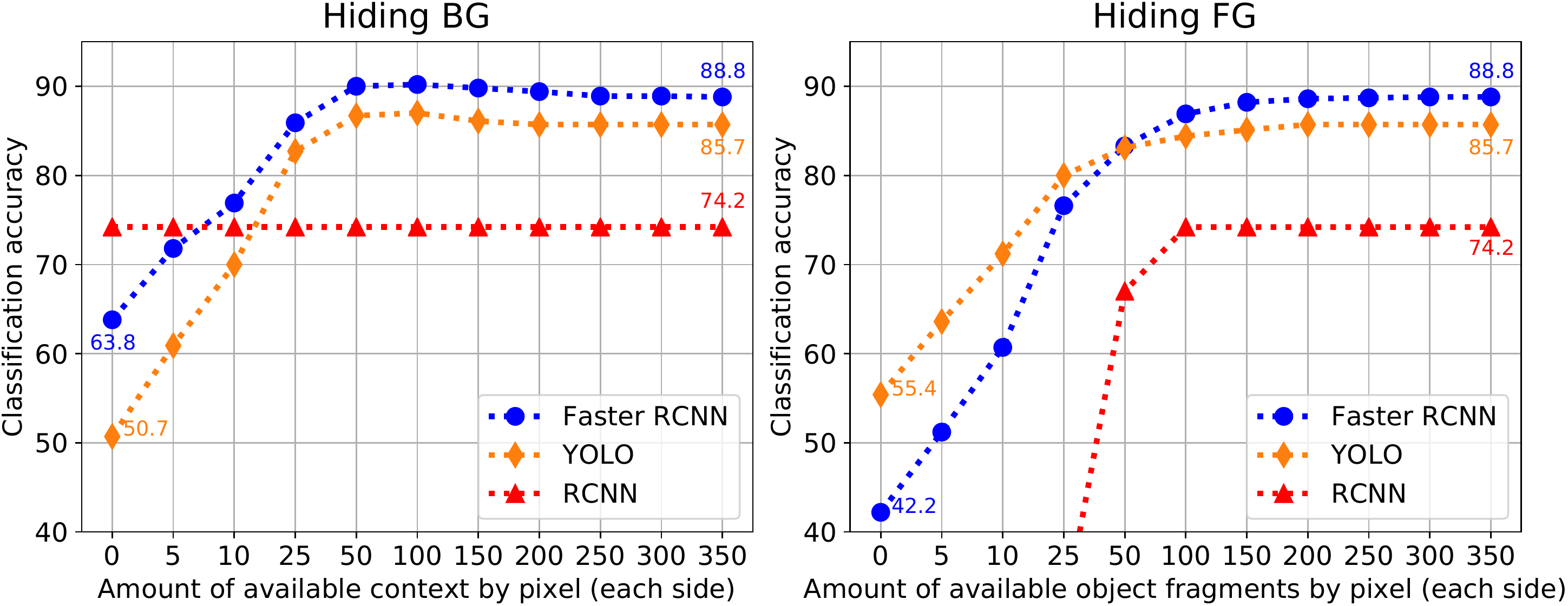}
	\caption{ Hiding Background and Foreground experiments on 80-class COCO with RCNN, Faster RCNN and YOLO. The x-axis shows how many pixels of context (Hiding BG) or object (Hiding FG) is available from each side. Both plots confirm Faster RCNN is less sensitive to context than YOLO. For Hiding BG, 50-100 pixels extra has best object-context correlation, which reduces when adding more background. For Hiding FG, R-CNN needs only 100 pixels to classify an object. }
	\label{fig:frcnn_bg}
\end{figure*}
To evaluate hiding BG and FG setups, we use classification accuracy. The effect of localization is minimized as following: For RCNN, ground truth box locations are used to crop the objects from the input image. For Faster RCNN, IoU threshold is set as 0.25.
For YOLO, the prediction is counted as correct If the center location of predicted and ground truth boxes for correct class label remain in the same grid cell.

\textbf{Hiding BG.} Results in~\fig{frcnn_bg} (left) confirms that the Crop-Input R-CNN does not depend on context. The Crop-FM Faster RCNN outperforms the NoCrop YOLO when no context is available. Both detectors have their peak when 100 pixels is added to each side of the object: $90.2\%$ for Faster RCNN and $87\%$ for YOLO. 
We hypothesize that the object-context correlation is best around 100 pixels and decreases as more background is taken into account.  

\begin{figure}[!h]
	\centering
	\includegraphics[width=0.85\linewidth]{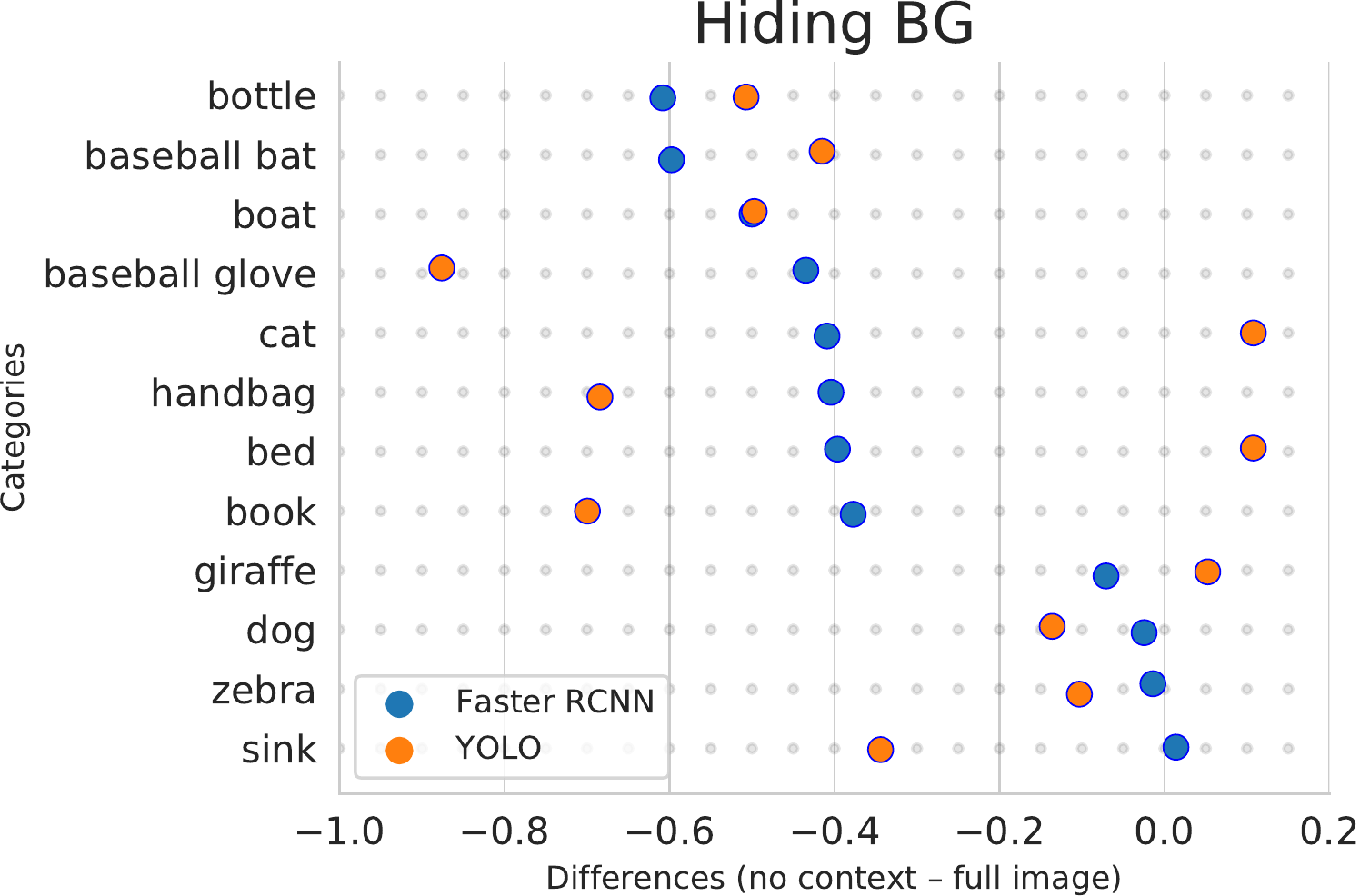}
	\includegraphics[width=0.85\linewidth]{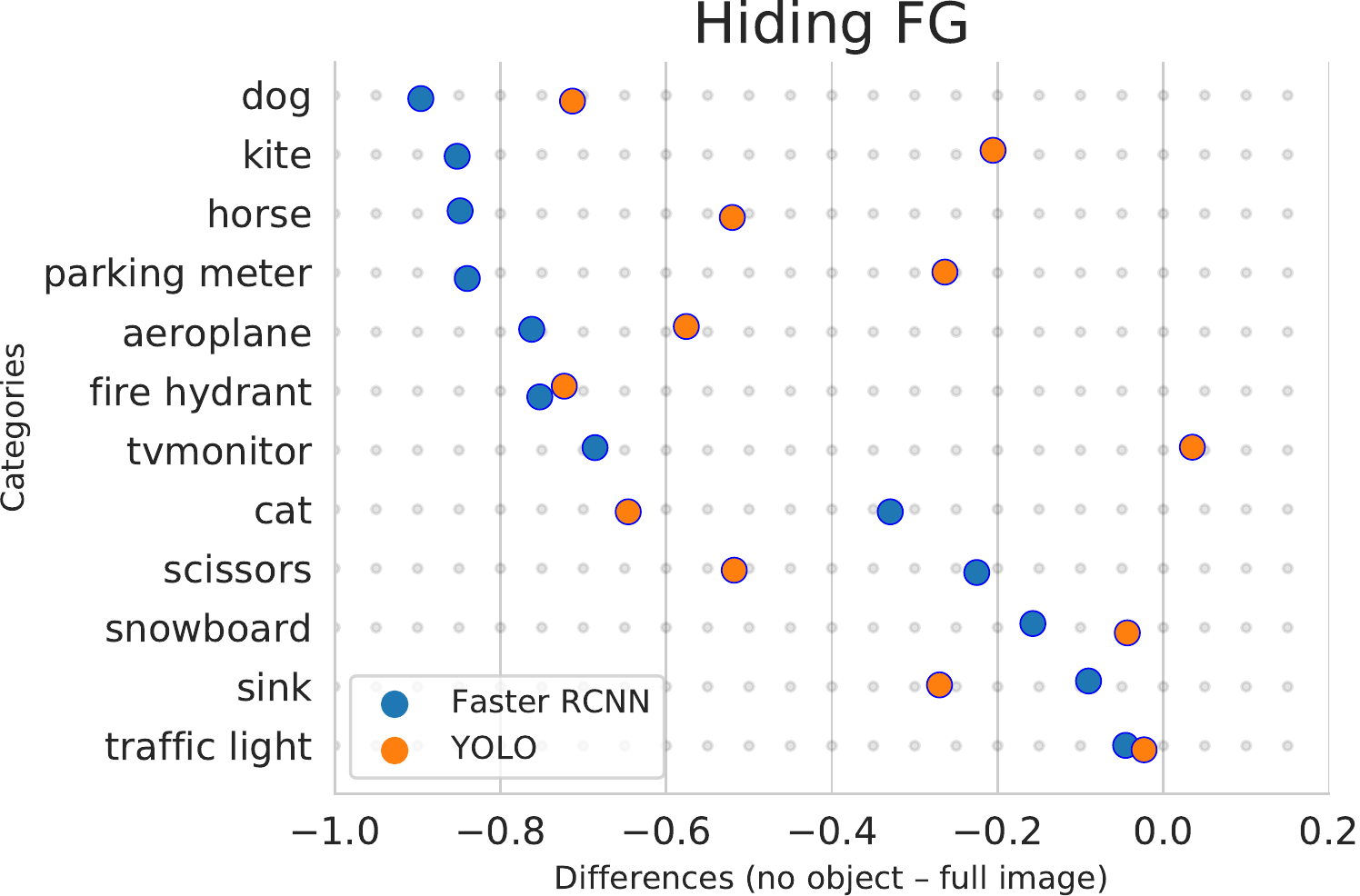}
	\caption{Evaluation of class-specific result of Faster RCNN and YOLO on COCO. Data points represent the 3 best and 3 worst-performing classes for each method according to difference between context/object and full image.}
	\label{fig:coco_cls_spe}
\end{figure}

For \textit{Hiding BG}, in~\fig{coco_cls_spe} (top) we show the classes that have the smallest  and the largest difference between no context and full context. For Faster RCNN, \textit{bottle} class is highly context dependent and loses 61\% without context. Besides, \textit{sink}  performs 2\% better without context.  For YOLO, \textit{baseball} \textit{glove}, \textit{handbag} and \textit{book}  lose more than $65\%$ due to lack of context. However, \textit{cat}, \textit{bed} and \textit{giraffe}  obtain better result when no context is used. 
Interestingly, 3/4 of classes obtains better result with using some amount of context rather than using full context. The fact also explains the performance increases in Fig. \ref{fig:frcnn_bg} (left) between 50 and 300 context ranges.

\textbf{Hiding FG.} Results in~\fig{frcnn_bg} (right) shows that YOLO is best in exploiting context when the full object is removed. Interestingly, the performance of Faster RCNN is still far from random for 80 classes without actually seeing the object while RCNN scores truly random with $\frac{1}{80} \approx 1.3\%$. RCNN can classify an object when more pixels are available and after 100 pixels, adding more pixels does not help.
Surprisingly,  when comparing \textit{Hiding FG} with \textit{Hiding BG} it shows that YOLO is $4.7\%$ better for not having the object when compared to not having the context. 

For \textit{Hiding FG}, in~\fig{coco_cls_spe} (bottom) we show the classes that have the smallest  and the largest difference between no object and full image.
Without seeing the actual object, Faster RCNN and YOLO can still classify a \textit{traffic light}. 
Classes like \textit{dog} and \textit{fire hydrant} lose more than $70\%$ performance for both methods when no object parts are visible.  
These classes  have high robustness to context change (Fig. \ref{fig:coco_cls_spe}-top), thus their object parts are crucial for their detection.
Surprisingly, \textit{tvmonitor} can be identified by YOLO 3.5\% better without seeing the object itself.

\section{Discussion and Conclusion}
In this paper, we investigate the effect of context on 3 different deep object detectors, (i) cropping the input (RCNN), (ii) partial context (Faster RCNN) and (iii) full context (YOLO). Experiments with Q-FMNIST and COCO datasets show that single and two stage methods access the context because of their large receptive fields excluding RCNN since it crops the ROIs from the input. Hiding BG and FG experiments indicate that context often improves the result until some extend and sometimes it degrades the performance. For YOLO, having no object visible outperforms having no context visible.

Generating realistic toy dataset for context experiments is challenging. Even if Q-FMNIST dataset is limited, it still provides controlled-context setup to compare common detectors. 
Besides, in hiding BG and FG experiments, object size matters and the effect of object size may supply insightful results, however, we focus on overall and class-specific performance indications rather than object sizes. 

\bibliographystyle{IEEEbib}
\bibliography{main}

\begin{thebibliography}{10}

\bibitem{gidaris2015object}
S.~Gidaris and N.~Komodakis,
\newblock ``Object detection via a multi-region and semantic segmentation-aware
  cnn model,''
\newblock in {\em ICCV}, 2015.

\bibitem{gupta2015exploring}
S.~Gupta, B.~Hariharan, and J.~Malik,
\newblock ``Exploring person context and local scene context for object
  detection,''
\newblock {\em arXiv preprint arXiv:1511.08177}, 2015.

\bibitem{Sun_2017_CVPR}
J.~Sun and D.~W. Jacobs,
\newblock ``Seeing what is not there: Learning context to determine where
  objects are missing,''
\newblock in {\em CVPR}, 2017.

\bibitem{Liu_2018}
Y.~Liu, R.~Wang, S.~S., and X.~Chen,
\newblock ``Structure inference net: Object detection using scene-level context
  and instance-level relationships,''
\newblock {\em CVPR}, 2018.

\bibitem{zhu2017couplenet}
Y.~Zhu, C.~Zhao, J.~Wang, X.~Zhao, Y.~Wu, and H.~Lu,
\newblock ``Couplenet: Coupling global structure with local parts for object
  detection,''
\newblock in {\em ICCV}, 2017.

\bibitem{fan2018object}
X.~Fan, H.~Guo, K.~Zheng, W.~Feng, and S.~Wang,
\newblock ``Object detection with mask-based feature encoding,''
\newblock {\em arXiv preprint arXiv:1802.03934}, 2018.

\bibitem{lim2019small}
J.-S. Lim, M.~Astrid, H.-J. Yoon, and S.-I. Lee,
\newblock ``Small object detection using context and attention,''
\newblock {\em arXiv preprint arXiv:1912.06319}, 2019.

\bibitem{bell2016inside}
S.~Bell, C~Lawrence~Zitnick, K.~Bala, and R.~Girshick,
\newblock ``Inside-outside net: Detecting objects in context with skip pooling
  and recurrent neural networks,''
\newblock in {\em CVPR}, 2016.

\bibitem{zhang2019bag}
Z.~Zhang, T.~He, H.~Zhang, Z.~Zhang, J.~Xie, and M.~Li,
\newblock ``Bag of freebies for training object detection neural networks,''
\newblock {\em arXiv preprint arXiv:1902.04103}, 2019.

\bibitem{singh2020don}
K.~K. Singh, D.~Mahajan, K.~Grauman, Y.~J. Lee, M.~Feiszli, and D.~Ghadiyaram,
\newblock ``Don't judge an object by its context: Learning to overcome
  contextual bias,''
\newblock in {\em CVPR}, 2020.

\bibitem{zhang2019putting}
M.~Zhang, C.~Tseng, and G.~Kreiman,
\newblock ``Putting visual object recognition in context,''
\newblock {\em arXiv preprint arXiv:1911.07349}, 2019.

\bibitem{Barnea_2019_CVPR}
E.~Barnea and O.~Ben-Shahar,
\newblock ``Exploring the bounds of the utility of context for object
  detection,''
\newblock in {\em CVPR}, 2019.

\bibitem{kayhan2021hallucination}
O.~S. Kayhan, B.~Vredebregt, and J.~C. van Gemert,
\newblock ``Hallucination in object detection — a study in visual part
  verification,''
\newblock in {\em ICIP}, 2021.

\bibitem{kayhan2020translation}
O.~S. Kayhan and J.~C.~van Gemert,
\newblock ``On translation invariance in {CNNs}: Convolutional layers can
  exploit absolute spatial location,''
\newblock in {\em CVPR}, 2020.

\bibitem{manfredi2020shift}
M.~Manfredi and Y.~Wang,
\newblock ``Shift equivariance in object detection,''
\newblock in {\em ECCV workshop}, 2020.

\bibitem{girshick2015region}
R.~Girshick, J.~Donahue, T.~Darrell, and J.~Malik,
\newblock ``Region-based convolutional networks for accurate object detection
  and segmentation,''
\newblock {\em PAMI}, 2015.

\bibitem{ren2015faster}
S.~Ren, K.~He, R.~Girshick, and J.~Sun,
\newblock ``Faster r-cnn: Towards real-time object detection with region
  proposal networks,''
\newblock in {\em Advances in neural information processing systems}, 2015, pp.
  91--99.

\bibitem{lin2017feature}
T.-Y. Lin, P.~Doll{\'a}r, R.~Girshick, K.~He, B.~Hariharan, and S.~Belongie,
\newblock ``Feature pyramid networks for object detection,''
\newblock in {\em CVPR}, 2017.

\bibitem{dai2017deformable}
J.~Dai, H.~Qi, Y.~Xiong, Y.~Li, G.~Zhang, H.~Hu, and Y.~Wei,
\newblock ``Deformable convolutional networks,''
\newblock in {\em ICCV}, 2017.

\bibitem{redmon2016you}
J.~Redmon, S.~Divvala, R.~Girshick, and A.~Farhadi,
\newblock ``You only look once: Unified, real-time object detection,''
\newblock in {\em Proceedings of the IEEE conference on computer vision and
  pattern recognition}, 2016.

\bibitem{Redmon_2017_CVPR}
J.~Redmon and A.~Farhadi,
\newblock ``Yolo9000: Better, faster, stronger,''
\newblock in {\em CVPR}, 2017.

\bibitem{redmon2018yolov3}
J.~Redmon and A.~Farhadi,
\newblock ``Yolov3: An incremental improvement,''
\newblock {\em arXiv preprint arXiv:1804.02767}, 2018.

\bibitem{liu2016ssd}
W.~Liu, D.~Anguelov, D.~Erhan, C.~Szegedy, S.~Reed, C.-Y. Fu, and A.~C. Berg,
\newblock ``Ssd: Single shot multibox detector,''
\newblock in {\em ECCV}, 2016.

\bibitem{lin2017focal}
T.-Y. Lin, P.~Goyal, R.~Girshick, K.~He, and P.~Doll{\'a}r,
\newblock ``Focal loss for dense object detection,''
\newblock in {\em ICCV}, 2017.

\bibitem{tan2019efficientdet}
M.~Tan, R.~Pang, and Q.~V. Le,
\newblock ``Efficientdet: Scalable and efficient object detection,''
\newblock {\em arXiv preprint arXiv:1911.09070}, 2019.

\bibitem{uijlings2013selective}
J.~R~.R. Uijlings, K.~E.~A. van~de Sande, T.~Gevers, and A.~W.~M. Smeulders,
\newblock ``Selective search for object recognition,''
\newblock {\em International journal of computer vision}, vol. 104, no. 2, pp.
  154--171, 2013.

\bibitem{xiao2017fashion}
H.~Xiao, K.~Rasul, and R.~Vollgraf,
\newblock ``Fashion-mnist: a novel image dataset for benchmarking machine
  learning algorithms,''
\newblock {\em arXiv preprint arXiv:1708.07747}, 2017.

\bibitem{he2017maskRCNN}
K.~He, G.~Gkioxari, P.~Doll{\'a}r, and R.~Girshick,
\newblock ``Mask r-cnn,''
\newblock in {\em ICCV}, 2017.

\bibitem{lin2014microsoft}
T.-Y. Lin, M.~Maire, S.~Belongie, J.~Hays, P.~Perona, D.~Ramanan,
  P.~Doll{\'a}r, and C.~L. Zitnick,
\newblock ``Microsoft coco: Common objects in context,''
\newblock in {\em ECCV}, 2014.

\end{thebibliography}

\end{document}